\documentclass[a4paper,twoside]{article}

\usepackage{booktabs}
\usepackage{graphicx}
\usepackage{amsfonts}
\usepackage{dsfont}
\usepackage{hyperref}
\usepackage{multirow}

\usepackage{epsfig}
\usepackage{subcaption}
\usepackage{calc}
\usepackage{amssymb}
\usepackage{amstext}
\usepackage{amsmath}
\usepackage{amsthm}
\usepackage{multicol}
\usepackage{pslatex}
\usepackage{algorithm2e}
\usepackage[bottom]{footmisc}
\usepackage{SCITEPRESS}     

\graphicspath{{media/}}     

\begin{document}

\title{Graph Memory: A Structured and Interpretable Framework for Modality-Agnostic Embedding-Based Inference}

\author{\authorname{Artur A. Oliveira\sup{1}\orcidAuthor{0000-0002-3606-1687}, Mateus Espadoto\sup{1}\orcidAuthor{0000-0002-1922-4309}, Roberto M. Cesar Jr.\sup{1}\orcidAuthor{0000-0003-2701-4288} and Roberto Hirata Jr.\sup{1}\orcidAuthor{0000-0003-3861-7260}}\affiliation{\sup{1}Institute of Mathematics and Statistics, University of São Paulo, Rua do Matão, 1010, São Paulo, Brazil}\email{\{arturao, mespadot, hirata\}@ime.usp.br, rmcesar@usp.br}}



\abstract{
We introduce \textbf{Graph Memory (GM)}, a structured non-parametric framework that represents an embedding space through a compact graph of reliability-annotated prototype regions. GM encodes local geometry and regional ambiguity through prototype relations and performs inference by diffusing query evidence across this structure, unifying instance retrieval, prototype-based reasoning, and graph diffusion within a single inductive and interpretable model. The framework is inherently modality-agnostic: in multimodal settings, independent prototype graphs are constructed for each modality and their calibrated predictions are combined through reliability-aware late fusion, enabling transparent integration of heterogeneous sources such as whole-slide images and gene-expression profiles. Experiments on synthetic benchmarks, breast histopathology (IDC), and the multimodal AURORA dataset show that GM matches or exceeds the accuracy of $k$NN and Label Spreading while providing substantially better calibration, smoother decision boundaries, and an order-of-magnitude smaller memory footprint. By explicitly modeling regional reliability and relational structure, GM offers a principled and interpretable approach to non-parametric inference across single- and multi-modal domains.
}
\onecolumn \maketitle \normalsize \setcounter{footnote}{0} \vfill

\section{\uppercase{Introduction}}

Modern embedding-based classifiers are typically organized around two dominant paradigms. Parametric models, such as linear-softmax heads, learn global decision boundaries in a fixed representation space, whereas non-parametric approaches such as prototypes or $k$NN ground predictions in local neighborhoods defined by geometric similarity. Although effective in practice, these models usually treat the embedding space as a collection of independent points or loosely formed clusters, with limited ability to capture how different regions relate, where uncertainty concentrates, or how local evidence should propagate across boundaries. Common summary statistics such as purity or silhouette describe within-region properties but overlook the broader relational geometry and multi-hop dependencies that naturally arise in embedding spaces.

We introduce \textbf{Graph Memory (GM)}, a structured and reusable representation that augments the embedding space with region-level prototypes and explicit relations among them. Each prototype summarizes a coherent region through its centroid and reliability attributes, such as purity, dispersion, and stability, while edges encode geometric proximity and persistent areas of ambiguity. The resulting reliability-weighted prototype graph provides a compact and interpretable view of the embedding manifold and enables context-aware inference that integrates local evidence with global structural cues.

GM operates in two stages. During memory construction, embeddings are clustered into prototype regions, annotated with reliability metadata, and connected through edges that reflect structural affinity. During inference, a query activates its most relevant prototypes, and prediction is obtained by diffusing this activation over the prototype graph in a way that respects both local similarity and regional reliability. This diffusion process yields calibrated and geometrically smooth decision functions while preserving interpretability, since the prototypes that accumulate the most evidence reveal the regions that support or challenge the prediction. GM therefore unifies instance retrieval, prototype-based reasoning, and graph diffusion within a single inductive framework, offering a principled alternative to density-sensitive methods that rely on full instance graphs.

A central appeal of GM is its ability to operate effectively in heterogeneous, sparsely sampled, or high-dimensional settings, conditions where traditional non-parametric methods struggle to capture global structure without incurring prohibitive memory or computational cost. These properties naturally extend to multimodal data, where observations from different modalities provide complementary geometric and semantic information. GM accommodates such heterogeneity by constructing an independent prototype graph for each modality and combining their predictions through reliability-aware late fusion, which preserves modality-specific interpretability while enabling principled cross-modal integration.

We evaluate GM across a spectrum of increasingly complex scenarios.
We show how retrieval, prototype abstraction, and graph diffusion can be expressed within GM unified model on controlled synthetic two-dimensional benchmarks,  and how the resulting decision functions exhibit smoothness and calibration advantages over classical $k$NN and Label Spreading~\cite{scikit-learn,sklearn_labelspreading_archive}. On high-dimensional breast histopathology (IDC)~\cite{janowczyk2016deep,cruz2014automatic}, GM achieves strong accuracy and competitive calibration using a compact bank of prototypes, demonstrating its data-efficiency and interpretability in real-world settings. Finally, on the multimodal AURORA dataset~\cite{GarciaRecio2023}, which combines whole-slide images with RNA-seq profiles of primary and metastatic breast tumors, GM exposes cross-modal agreement and disagreement, improves calibration through reliability-aware fusion, and remains robust in an extremely small-sample regime.

Overall, Graph Memory provides a compact, interpretable, and modality-agnostic structure for embedding-based inference. It supports smooth decision boundaries, calibrated predictions, and principled multimodal integration within a unified inductive framework, validated across synthetic benchmarks, real-world image data, and challenging clinical multimodal applications.

\section{\uppercase{Related Work}}

This section surveys prior art relevant to embedding-space inference, organized from general to specific. We first contrast parametric and non-parametric classifiers, then review prototype and cluster-based methods, instance retrieval and semi-parametric adapters, and graph-based semi-supervised learning. We next discuss structured relations beyond pure distance (hierarchies and temporal dynamics), summarize efficiency results for serving large non-parametric memories, and close with interpretability, calibration, and cluster-quality assessment. Throughout, we emphasize where existing lines fall short, most notably in lacking a persistent, region-level representation that encodes reliability, ambiguity, and relations among regions in a reusable form.

\paragraph{Parametric \emph{vs.} non-parametric classifiers.}
Classification based on embeddings is commonly realized with \emph{parametric} heads (e.g., linear-softmax layers) that learn global decision boundaries, or with \emph{non-parametric} schemes (e.g., prototypes, $k$NN) that rely on neighborhood evidence. While effective, both families rarely maintain an explicit, reusable representation of how \emph{regions} of the space relate, where ambiguity persists, which areas are reliably separated, and how evidence should flow across regions.

\paragraph{Prototype and cluster-based classifiers.}
Nearest-class-mean and related prototype methods summarize classes via one or a few representatives, while few-shot variants such as Prototypical Networks learn embeddings where class means become discriminative \cite{snell2017prototypical}. Prototype-part models like ProtoPNet provide ``this-looks-like-that'' interpretability within parametric networks \cite{chen2019protopnet}. However, prototypes are typically treated in isolation: relations among regions (e.g., persistent cross-class ambiguity, reliability differences between prototypes) are not explicitly modeled, limiting context-aware inference.

\paragraph{Instance retrieval and semi-parametric adapters.}

Instance-based inference spans classical $k$NN to Deep $k$NN (DkNN) \cite{papernot2018dknn}, which retrieves neighbors across multiple network layers to improve robustness and interpretability. Subsequent semi-parametric adapters integrate retrieval with pretrained encoders at scale, e.g., $k$NN with frozen vision backbones for privacy-aware continual learning and deletion \cite{doerrich2024privacyknn}, class-conditional memory that augments parametric logits \cite{bhardwaj2023knncm}, and kNN-Adapter for black-box LM domain adaptation \cite{huang2023knnadapter}. While these approaches underscore the value of non-parametric memory, they operate at the \emph{instance} or \emph{activation} level and are tightly coupled to specific deep architectures. In contrast, embedding-level or prototype-level representations can serve as a universal non-parametric head independent of the encoder.

\paragraph{Graph-based semi-supervised learning (Graph SSL).}

Graph SSL constructs instance graphs and propagates labels harmonically~\cite{zhu2003ssl,belkin2006manifold}.
Graph Convolutional Networks (GCNs)~\cite{kipf2017gcn} extend this idea by learning parametric message-passing rules over the same graphs.
Work in this line of research primarily focuses on designing propagation operators given a fixed instance graph, typically in transductive settings where test samples must be inserted into the graph before inference.

An influential variant of graph diffusion is \emph{Personalized PageRank} (PPR)~\cite{haveliwala2002topic}, which performs a random-walk-with-restart biased toward a user-defined seed distribution.  
PPR exemplifies how restart-based diffusion preserves personalization while exploiting global graph structure

Open questions remain around \emph{graph construction}: which nodes to represent (instances vs.\ prototypes), which edge semantics to encode, and how to attach new queries in an inductive and interpretable manner.  
Zhou \emph{et al.}~\cite{zhou2004llgc} showed that the steady-state solution of a linear propagation process yields a closed-form classifier, linking manifold flatness to global label consistency.

\paragraph{Structured relations beyond distances.}

Beyond isotropic similarity, hierarchical relations have been modeled with partial-order and box embeddings.  

Objective design and negative sampling are crucial for reconstructing high-fidelity taxonomies \cite{taxo2020boxes}.  

These approaches suggest richer edge semantics (e.g., containment, overlap) but do not yield a reusable region-level memory for inductive classification.

\paragraph{Temporal relations in graphs.}

Time-decayed line graphs formalize dynamic adjacency with explicit decay kernels \cite{tdlg2022timedecayed}.  

Such models provide a template for injecting recency and historical ambiguity into edge weights, though they do not address prototype construction or region-level reliability.

\paragraph{Efficiency and scalability of non-parametric memories.}

Dhulipala \emph{et al.} propose MUVERA (Multi-Vector Retrieval via Fixed Dimensional Encodings)~\cite{muvera2024fde}, a retrieval mechanism that compresses multi-vector interaction into single-vector maximum inner-product search with high recall.  

This shows that large non-parametric memories can be served efficiently, but these works do not specify what structural or reliability information such memories should encode.

\paragraph{Interpretability, calibration, and cluster quality.}

Prototype-based interpretability \cite{chen2019protopnet} complements post-hoc calibration techniques such as temperature scaling \cite{guo2017calibration}, Dirichlet calibration \cite{kull2019dirichlet}, and energy-based OOD (out-of-distribution) scoring \cite{liu2020energy}.  
Orthogonally, clustering indices such as Silhouette~\cite{rousseeuw1987silhouettes}, Davies–Bouldin~\cite{4766909}, Dunn~\cite{Dunn1974WellSeparatedCA}, and Calinski–Harabasz~\cite{calinskiharabasz} evaluate intra- and inter-cluster structure.  
Liu \emph{et al.}~\cite{liu2010understanding} show these indices capture complementary aspects, motivating multi-metric region-quality assessment.

\paragraph{Summary of gaps.}
Prior work contributes essential ingredients: prototypes for interpretability, retrieval for adaptability, semi-parametric adapters for scalability, graph SSL for relational reasoning, hierarchical and temporal modeling for richer relations, and efficient neural-network backends for serving. Taken in isolation, however, these lines often (i) treat prototypes or instances without encoding inter-region relations, (ii) lack a persistent, reusable memory over regions, (iii) omit reliability and ambiguity as first-class signals, or (iv) leave graph construction under-specified for scalable, query-time attachment. This leaves open the need for a structured, prototype-level graph memory that jointly encodes reliability, ambiguity, and relations among regions, while remaining compatible with modern retrieval and inference mechanisms, and applicable beyond deep architectures.

\section{\uppercase{Method}}
\label{sec:method}

\noindent
\textbf{Graph Memory (GM)} is a reusable non-parametric structure that represents an embedding space as a graph of prototype regions.
Furthermore, GM is \emph{embedding-agnostic} and attaches to any fixed encoder without retraining or architectural changes, acting as a flexible non-parametric head usable in unimodal and multimodal settings.
Each prototype summarizes a local neighborhood of the manifold, and edges encode geometric affinity, ambiguity, and contextual relations between regions. Unlike nearest-distance methods such as $k$NN~\cite{knn}, GM performs \emph{evidence diffusion}: the query defines a starting distribution over prototypes, and a random-walk–with–restart propagates this evidence across the reliability-weighted graph. This is closely related to Personalized PageRank~\cite{haveliwala2002topic}, where diffusion is anchored at a personalized seed; here, the personalization vector is the query-dependent activation $z_0$. The process converges to a \emph{stationary distribution} whose components correspond to the long-run visitation frequencies of this walk, integrating local similarity with global structural cues and preserving the query-dependent nature of the inference.

\subsection{Overview}
Given a labeled embedding dataset $\mathcal{D}=\{(x_i,y_i)\}_{i=1}^{N}$, where $x_i=f(x_i^{\mathrm{raw}})\in\mathbb{R}^d$ are fixed embeddings and $y_i\in\{1,\dots,C\}$ are class labels, we construct a graph $\mathcal{G}=(\mathcal{V},\mathcal{E})$ whose nodes $\mathcal{V}=\{1,\dots,K\}$ are cluster-level prototypes. Each prototype summarizes a subset $S_c\subseteq\mathcal{D}$ and edges $(c,c')\in\mathcal{E}$ encode relational affinity. The pipeline has three parts: (1) joint prototype construction, (2) relation graph formation, and (3) inductive inference by graph diffusion.

\subsection{Prototype Selection}

We partition the embeddings $\{x_i\}$ into $K$ groups using $K$-means or $K$-medoids across all classes jointly, preserving mixed boundary regions that per-class clustering would treat separately. 
Each prototype $c$ is represented by its centroid
\[
\mu_c = \frac{1}{|S_c|}\sum_{x_i\in S_c} x_i,
\]
and the following attributes are stored:
\begin{itemize}
  \item \textbf{Support:} $|S_c|$, number of samples represented by prototype $c$.
  \item \textbf{Dominant class:} $y_c=\arg\max_y |\{x_i\in S_c:y_i=y\}|$, the most frequent class within $S_c$.
  \item \textbf{Purity:} $\pi_c=\frac{\max_y |\{x_i\in S_c:y_i=y\}|}{|S_c|}$, the proportion of samples belonging to the dominant class, used later as a reliability factor.
\end{itemize}
All statistics are normalized by the local population $|S_c|$, ensuring that prototype attributes reflect the stability of the region they summarize.

\subsection{Prototype Quality and Reliability}

Each prototype is further characterized by geometric and stability metrics that are combined into a scalar reliability score $r_c\in[0,1]$.\\

\noindent\textbf{(1) Normalized silhouette.}
For each sample $x_i\!\in\!S_c$, let
\begin{align}
a(i) &= \tfrac{1}{|S_c|-1}\sum_{x_j\in S_c,\,j\ne i}\|x_i-x_j\|_2, 
&\text{(intra-cluster dist.)}\\
b(i) &= \min_{c'\ne c}\tfrac{1}{|S_{c'}|}\sum_{x_k\in S_{c'}}\|x_i-x_k\|_2,
&\text{(nearest-cluster dist.)}
\end{align}
and compute the normalized silhouette value
\begin{align*}
s_c = \frac{1}{|S_c|}\sum_{x_i\in S_c}\frac{\text{sil}(i)+1}{2}, 
\quad\\\text{where }\text{sil}(i)=\frac{b(i)-a(i)}{\max\{a(i),b(i)\}}.
\end{align*}
This rescales the original silhouette from $[-1,1]$ to $[0,1]$, aligning its range with the other metrics.\\

\noindent\textbf{(2) Dispersion.}
The internal variance relative to the centroid:
\[
v_c = \tfrac{1}{|S_c|}\sum_{x_i\in S_c}\|x_i-\mu_c\|_2^2.
\]
Low $v_c$ indicates compact geometry, while high values imply intra-class spread or outliers.\\

\noindent\textbf{(3) Margin.}
Separation from prototypes of different dominant classes:
\[
m_c = \min_{c':\,y_{c'}\neq y_c}\|\mu_c - \mu_{c'}\|_2.
\]
\\
\noindent\textbf{(4) Instability.}
Sensitivity to embedding perturbations. 
We sample $x_i^\delta = x_i + \delta_i$ (small Gaussian noise) and measure reassignment rate:
\[
\rho_c = \tfrac{1}{|S_c|}\sum_{x_i\in S_c}\mathbf{1}\!\left[\operatorname{NN}(x_i^\delta)\neq c\right].
\]
Lower $\rho_c$ means more stable assignments.\\

\noindent\textbf{(5) Robust normalization.}
Among the prototype metrics, only the margin $m_c$ and dispersion $v_c$ are unbounded and directly depend on the scale of the embedding space.
To make them comparable with the bounded quantities ($s_c$, $\rho_c$, $\pi_c$), we apply a robust rescaling based on median and interquartile range:
\begin{equation}
\bar{x}_c = \sigma\!\left(\frac{x_c - \mathrm{med}(x)}{\mathrm{IQR}(x)}\right)\label{eq:iqr}
\end{equation}
\begin{equation*}
\mathrm{IQR}(x)=Q_{0.75}(x)-Q_{0.25}(x)    
\end{equation*}
where $\sigma(\cdot)$ is the logistic function.  
This transformation, akin to robust scaling \cite{rouseeuw1993alternatives}, compresses extreme values and removes sensitivity to global scale, ensuring that large distances or variances do not dominate the composite reliability.  
For metrics already bounded in $[0,1]$ (e.g., $s_c$, $\rho_c$, $\pi_c$), no additional normalization is applied.\\

\noindent\textbf{(6) Composite reliability.}
The overall reliability is then
\[
r_c = \sigma\!\Big(
\lambda_1\,s_c
+ \lambda_2\,\bar{m}_c
+ \lambda_5\,\pi_c
- \lambda_3\,\rho_c
- \lambda_4\,\bar{v}_c
\Big),
\]
where $\lambda_i$ are non-negative weights (default $=1$). 
Bounded metrics ($s_c$, $\rho_c$, $\pi_c$) are used directly, while unbounded metrics ($m_c$, $v_c$) are robustly normalized as in Eq.~\ref{eq:iqr}. 
This aggregation balances compactness, separation, stability, and purity on a comparable numerical scale.

\subsection{Graph Memory Construction}

Prototype relations are encoded in a weighted graph built in the embedding space, where edges reflect both geometric affinity and contextual reliability between regions.
Each edge weight measures the local affinity between prototypes $c$ and $c'$ through a Gaussian kernel:
\[
A_{cc'} =
\begin{cases}
\exp(-\beta\,\|\mu_c - \mu_{c'}\|_2^2), & c' \in \mathrm{KNN}_k(c),\\[4pt]
0, & \text{otherwise.}
\end{cases}
\]
The adjacency matrix $A$ is symmetrized and row-normalized to obtain a stochastic transition matrix
\[
S = D^{-1}A, \qquad D_{cc} = \sum_{c'} A_{cc'},
\]
which defines the propagation operator used during inference.  
This construction captures the local geometric structure of the prototype manifold and provides the topology for evidence diffusion.
In practice, these edges are primarily used to quantify contextual coherence, reliability propagation, and regional ambiguity, rather than to modify class boundaries directly.

\subsection{Inference by Graph Diffusion}

Given a query embedding $x_q$, GM first computes an initial activation over prototypes,
\[
z_0(c) = \exp(-\beta\|x_q - \mu_c\|_2^2)\, r_c,
\]
where $\mu_c$ is the prototype centroid and $r_c$ its reliability.  
This $z_0$ serves as a \emph{starting distribution}, indicating how strongly the query is associated with each prototype.

Instead of predicting directly from these local affinities, GM performs \emph{evidence diffusion} on the prototype graph.  
The activation is propagated by a random-walk--with-restart update,
\[
z^{(t+1)} = (1-\alpha)\,z_0 + \alpha\, S z^{(t)},
\]
where $S$ is the row-normalized affinity matrix and $\alpha\!\in\![0,1)$ controls how much evidence flows through the graph at each step.  
With probability $\alpha$ the walk transitions according to $S$, and with probability $1-\alpha$ it restarts from $z_0$.  
This process is guaranteed to converge to a unique stationary activation vector
\begin{equation}
z = (I - \alpha S)^{-1} z_0,
\label{eq:diffusion_closed_form}
\end{equation}
which corresponds to the steady state of the classical label-propagation formulation of Zhou \emph{et al.}~\cite{zhou2004llgc}.  
The components of $z$ reflect the long-run visitation frequencies of this walk anchored at the query, blending local similarity with global structural context.

Class probabilities are obtained by aggregating stationary activations over prototypes with the same dominant label:
\[
p(y \mid x_q) \propto \sum_{c:\,y_c = y} z_c,
\qquad
\hat{y}_q = \arg\max_y p(y \mid x_q).
\]

When $\alpha = 0$, GM reduces to a purely local scheme analogous to weighted $k$NN.  
As $\alpha$ increases, diffusion incorporates more relational information from the prototype graph, yielding smoother decision boundaries and more calibrated predictions.

\subsection{Interpretation and Comparison to Baselines}
\label{sec:baselines comparison}

Graph Memory (GM) unifies several classical non-parametric inference schemes within a single framework.  
Depending on the configuration of prototypes, diffusion strength $\alpha$, and reliability terms $r_c$, GM recovers well-known methods as limiting cases:

\begin{itemize}
    \item \textbf{$k$NN:} when each prototype represents a single instance and diffusion is disabled ($\alpha{=}0$), GM reduces to standard $k$-nearest-neighbor classification.
    \item \textbf{Nearest-centroid:} when all samples of each class collapse to a single prototype and $r_c\!\equiv\!1$, inference corresponds to a nearest-class-mean classifier.
    \item \textbf{Label propagation:} when diffusion is active ($\alpha{>}0$) and reliabilities are uniform, the update rule becomes the steady-state solution of a label-propagation system, yielding a semi-parametric graph-based variant.
\end{itemize}

\noindent
Unlike these baselines, GM operates on a compact set of region-level prototypes that encode both local geometry and reliability, combining the interpretability of prototype models with the contextual smoothing of graph diffusion.  
This enables \textbf{calibrated, context-aware, and geometrically regular} predictions while maintaining the simplicity and scalability of non-parametric inference.  
Because it is embedding-agnostic, GM can be applied as a non-parametric head on top of any pre-trained encoder, from deep neural networks to handcrafted or multimodal feature spaces.  
By construction, GM provides a principled bridge between instance-level, class-level, and graph-based reasoning, generalizing $k$NN, nearest-centroid, and label-propagation methods under a single relational formulation.
Furthermore, GM offers localized, prototype-level explanations that reveal which regions support a given decision and how reliable their contributions are, extending prior work on dimensionality reduction~\cite{oliveira2021improving}, supervised decision-boundary visualization~\cite{oliveira2022sdbm}, and stability analysis of decision maps~\cite{oliveira2023stability}.

\paragraph{Role of edges.}
While prototype activations alone often suffice for accurate predictions, the edge structure provides a complementary view of the embedding manifold.
Connectivity patterns among prototypes reveal where regions are coherent, ambiguous, or sparsely supported, which enables outlier detection and visual explanations of class overlap.
Thus, GM distinguishes between local evidence (nodes) and contextual reliability (edges), yielding an interpretable non-parametric memory that can explain and calibrate predictions.

\subsection{Multimodal Inference with Independent Graph Memories}
\label{subsec:multimodal}

The Graph Memory (GM) formulation is inherently modality-agnostic and can operate on embeddings produced by any fixed encoder. In multimodal settings, samples are represented in heterogeneous feature spaces, such as images, genomic profiles, or structured clinical data. Direct early fusion of these modalities is often brittle, since differences in scale, noise characteristics, and sampling density can obscure modality-specific structure and reduce interpretability.

GM accommodates multimodal data by constructing \textbf{independent graph memories}, one per modality, each capturing the geometric organization, reliability patterns, and region-level interactions within its own embedding manifold. These memories operate in parallel: each modality produces a calibrated probability estimate through its own prototype graph and diffusion process. Their outputs are subsequently combined using reliability-aware late fusion, allowing complementary evidence to be integrated without requiring early fusion or cross-modal alignment.

\paragraph{Independent memories.}
Let $\mathcal{X}=\{x^{(1)},\dots,x^{(M)}\}$ denote $M$ aligned modalities describing the same set of $N$ samples, and let $f_m(\cdot)$ be the fixed encoder for modality $m$.  
Each encoder produces embeddings $h_i^{(m)} = f_m(x_i^{(m)}) \in \mathbb{R}^{d_m}$.  
For every modality, we construct an independent graph memory
\[
\mathrm{GM}_m = (\mathcal{P}_m, S_m, r_m),
\]
where $\mathcal{P}_m = \{\mu_c^{(m)}\}_{c=1}^{K_m}$ are prototype centroids, $S_m$ is the row-stochastic diffusion matrix derived from pairwise affinities among prototypes, and $r_m$ are reliability weights computed as described in Sec.~\ref{sec:method}.  
Each $\mathrm{GM}_m$ operates on its own embedding space, encoding how local regions relate and how reliable they are.  
Inference for a query sample $x_q^{(m)}$ proceeds independently within each modality:
\[
p_m(y \mid x_q^{(m)})
\;\propto\;
\sum_{c:\,y_c^{(m)}=y} z^{(m)}_c,
\qquad
z^{(m)} = (I - \alpha S_m)^{-1} z^{(m)}_0,
\]
where the initial activation vector is
\[
z^{(m)}_0(c) = \exp\!\big(-\beta_m \|h_q^{(m)} - \mu_c^{(m)}\|_2^2\big)\,r_c^{(m)}.
\]
Thus, each modality yields an independent, reliability-weighted posterior $p_m(y)$ that reflects its own geometric structure and evidence distribution.

\paragraph{Interpretable consensus.}
Rather than enforcing early fusion in the embedding space, GM retains separate posteriors $\{p_m(y)\}_{m=1}^{M}$ that can be compared or combined at the decision level.  
A reliability-aware consensus prediction is obtained as
\begin{equation}
p_{\mathrm{fused}}(y)
=
\frac{\sum_m \omega_m\,r_m^{(q)}\,p_m(y)}{\sum_m \omega_m\,r_m^{(q)}},
\label{eq:fusion_rule}
\end{equation}
\noindent
where $r_m^{(q)}$ denotes the mean reliability of prototypes activated for query $q$ in modality $m$, and $\omega_m$ is an optional scalar \emph{modality weight} controlling the relative influence of each modality.  
If no prior preference is given, $\omega_m{=}1$ for all $m$, yielding an unweighted reliability-based fusion.  
This late-fusion formulation preserves interpretability by keeping per-modality reasoning explicit, while still enabling a unified consensus prediction when desired.

\paragraph{Cross-modal agreement.}
Disagreement between modalities is informative rather than undesirable.  
For any pair of modalities $(A,B)$, we define a simple consistency index
\[
A_{AB} = 1 - \tfrac{1}{2}\sum_y |p_A(y) - p_B(y)|,
\]
where low $A_{AB}$ indicates high disagreement and potential uncertainty.  
Tracking agreement across modalities exposes complementary or conflicting evidence, enabling analyses of uncertainty, redundancy, and complementary signal across feature spaces.

\paragraph{Benefits.}
This modular multimodal formulation preserves the key strengths of GM, compactness, diffusion-based reasoning, and localized interpretability, while avoiding the pitfalls of early fusion.
Each memory remains fully interpretable and calibrated within its own modality, yet the system as a whole provides a coherent framework for comparing, combining, or auditing evidence across modalities.  
By treating cross-modal discordance as an explicit output rather than a modeling failure, GM multimodal inference is transparent and trustworthy in a broad range of applications.

\section{\uppercase{Experimental Evaluation}}

We evaluate Graph Memory (GM) on synthetic benchmarks, breast histopathology (IDC), and multimodal biomedical data (AURORA), focusing on accuracy, calibration, smoothness, and interpretability. All methods operate on the same frozen embeddings, train/test splits, and preprocessing. Hyperparameters are selected on validation data.
Our experimental code is available at \url{https://github.com/arturandre/clip-qr-kg}

\paragraph{Metrics.}
We report Top-1 accuracy, negative log-likelihood (NLL), and the smoothness of decision functions via Dirichlet energy.
All results are reported as `mean\ (std)' over ten runs.
For 2D datasets, smoothness is estimated from the mean squared gradient of the class-probability field:
\[
E_{2\text{D}}
= \mathbb{E}_{(x,y)\in\mathcal{G}}
\!\left[\|\nabla p(y{=}1\mid x)\|_2^2\right],
\]
where lower values indicate smoother and more stable decision transitions.
For higher-dimensional embeddings, the same principle is measured via the \emph{graph Dirichlet energy}, a standard notion of smoothness in signal processing on graphs~\cite{shuman2013signal}:
\begin{equation}
\label{eq:crazy-grad-high-d}
E(f) \;=\; \tfrac{1}{2|\mathcal{E}|}\!\sum_{(i,j)\in \mathcal{E}} w_{ij}\,\bigl(f_i - f_j\bigr)^2
\end{equation}
\begin{align*}
w_{ij} = \exp(-\beta\,\|x_i - x_j\|_2^2)
\end{align*}
where $f_i = p(y{=}1\!\mid\!x_i)$ and $\mathcal{E}$ denotes $k$-nearest-neighbor edges.
Lower energy values correspond to smoother, more regular decision functions across the embedding manifold.

\subsection{Baselines}
We compare GM to representative parametric and non-parametric models:
\textbf{Linear probe:} multinomial logistic regression,
\textbf{$k$NN:} instance-level $k$-nearest neighbors,
\textbf{Budget-$k$NN:} $k$NN trained on a stratified subset whose size matches GM's prototype budget ($|\text{train}|{=}K$),
and \textbf{Label Spreading (LS):} scikit-learn transductive label spreading on an instance graph with $n$-neighbors~\cite{scikit-learn,sklearn_labelspreading_archive}.

\subsection{Synthetic datasets}
\label{subsec:synthetic}

\begin{figure*}[h]
\centering
\includegraphics[width=0.8\linewidth]{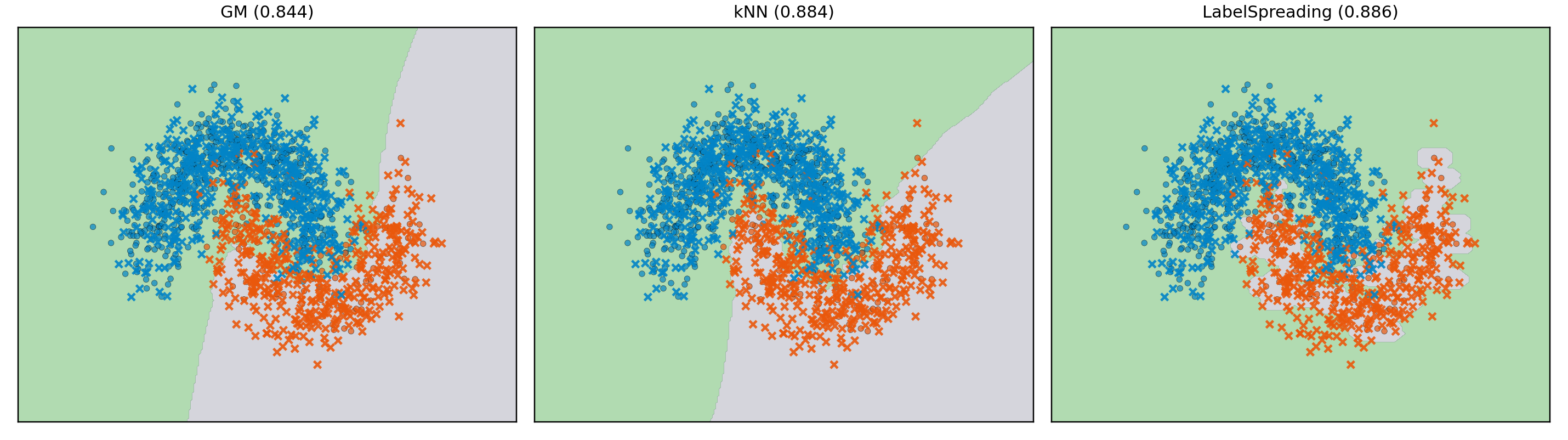}
\caption{Decision regions on synthetic data. GM yields flatter, reliability-weighted boundaries while retaining prototype-level interpretability.}
\label{fig:toy}
\end{figure*}

\begin{figure*}[h]
\centering
\includegraphics[width=0.8\linewidth]{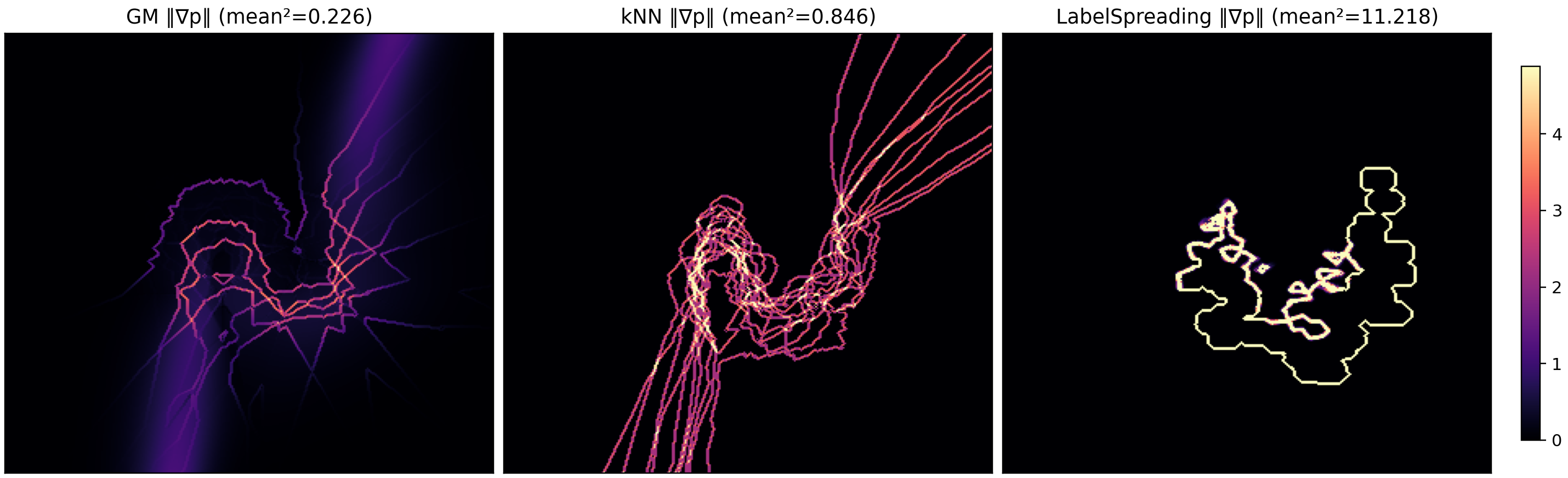}
\caption{
\textbf{Dirichlet energy maps} (visualized as gradient-magnitude fields, $\|\nabla p(y{=}1\mid x)\|$).  
Lower values indicate smoother and more stable decision transitions.  
GM produces lower-energy, reliability-weighted boundaries and avoids the sample-hugging artifacts observed in Label Spreading and $k$NN, particularly near class interfaces.
}
\label{fig:smooth}
\end{figure*}

We use the \textit{moons} and \textit{circles} datasets~\cite{scikit-learn} (4000 samples each, 50/50 split).
Class-imbalance experiments downsample the minority class to an $8{:}1$ ratio.
GM uses $K=120$ prototypes (balanced) or $K=112$ (imbalanced) and $k_{\text{graph}}{=}10$, $\alpha{=}0.5$, $\beta{=}0.1$, and attach-$k{=}8$.
Figures~\ref{fig:toy} and \ref{fig:smooth} show decision regions and gradient maps.

Table~\ref{tab:synth_acc_nll} shows that GM matches $k$NN and LS in accuracy but achieves consistently lower NLL, using only $\sim10\%$ of training samples.
GM remains fully \emph{inductive}, unlike LS, and robust under long-tail imbalance.
Budget-$k$NN highlights that naive compression severely hurts performance, whereas GM remains stable due to structured region prototypes.
Table~\ref{tab:smooth} reports smoothness: GM reduces Dirichlet energy by \textbf{$30\sim40$\%} relative to $k$NN and by \textbf{94\%} relative to LS, confirming flatter, more reliable boundaries.

\begin{table}[]
\centering
\resizebox{\columnwidth}{!}{%
\begin{tabular}{l|l|lcc}
\hline
Dataset                   & Imbal.?                                 & Method            & Acc $\uparrow$         & NLL $\downarrow$       \\ \hline
\multirow{10}{*}{moons}   & \multirow{5}{*}{No}                        & GM (P=120) \emph{Ours}       & 0.936 (0.008)          & \textbf{0.178 (0.010)} \\
&                                            & kNN-budget(P=120) & 0.935 (0.006)          & 0.298 (0.077)          \\ 
&                                            & kNN               & \textbf{0.940 (0.007)} & 0.402 (0.062)          \\
&                                            & Label Spreading    & 0.936 (0.010)          & 0.226 (0.018)          \\
                          &                                            & Linear            & 0.857 (0.005)          & 0.318 (0.006)          \\\cline{2-5} 
                          & \multirow{5}{*}{Yes} & GM (P=112)  \emph{Ours}       & 0.871 (0.018)          & \textbf{0.339 (0.052)} \\
                          &                                            & kNN-budget(P=112) & 0.689 (0.033)          & 1.153 (0.414)          \\ 
                          &                                            & kNN               & 0.880 (0.012)          & 0.758 (0.192)          \\
                          &                                            & Label Spreading    & \textbf{0.889 (0.014)} & 0.526 (0.065)          \\
                          &                                            & Linear            & 0.780 (0.011)          & 0.531 (0.026)          \\\hline
\multirow{10}{*}{circles} & \multirow{5}{*}{No}                        & GM (P=120)  \emph{Ours}       & \textbf{0.947 (0.005)} & \textbf{0.178 (0.009)} \\
&                                            & kNN-budget(P=120) & 0.936 (0.007)          & 0.353 (0.024)          \\
&                                            & kNN               & \textbf{0.947 (0.004)} & 0.331 (0.043)          \\
&                                            & Label Spreading    & 0.940 (0.006)          & 0.206 (0.028)          \\
                          &                                            & Linear            & 0.499 (0.008)          & 0.693 (0.000)          \\\cline{2-5} 
                          & \multirow{5}{*}{Yes} & GM (P=112)  \emph{Ours}       & 0.859 (0.014)          & \textbf{0.327 (0.013)} \\
                          &                                            & kNN-budget(P=112) & 0.857 (0.015)          & 0.543 (0.132)          \\
                          &                                            & kNN               & \textbf{0.892 (0.008)} & 0.654 (0.041)          \\
                          &                                            & Label Spreading    & 0.883 (0.007)          & 0.472 (0.018)          \\
                          &                                            & Linear            & 0.500 (0.000)          & 1.158 (0.000)          \\\hline
\end{tabular}
}
\caption{
\textbf{Accuracy and calibration on synthetic binary datasets.}
GM matches $k$NN and Label Spreading in accuracy while achieving better calibration and requiring only $\sim$10\% of the samples.
Unlike $k$NN and Label Spreading, GM remains inductive and robust under the $8{:}1$ long-tail regime.
}
\label{tab:synth_acc_nll}
\end{table}

\begin{table}[]
\centering
\resizebox{\columnwidth}{!}{%
\begin{tabular}{l|l|lc}
\hline
Dataset       & Imbal.?        & Method         & $E_{2\text{D}} \downarrow$    \\ \hline
\multirow{6}{*}{moons}   & \multirow{3}{*}{No}      & \small{GM (P=120)}     & \textbf{\small{0.927 (0.048)}} \\
& & \small{kNN} & \small{1.410 (0.095)}          \\  
& & \small{Label Spreading} & \small{17.705 (1.322)}         \\\cline{2-4}
& \multirow{3}{*}{Yes} & \small{GM (P=112)}     & \textbf{\small{0.834 (0.054)}} \\
& & \small{kNN} & \small{1.177 (0.066)}          \\ 
& & \small{Label Spreading} & \small{14.216 (1.275)}         \\\hline
\multirow{6}{*}{circles} & \multirow{3}{*}{No}      & \small{GM (P=120)}     & \textbf{\small{0.723 (0.063)}} \\
& & \small{kNN} & \small{1.065 (0.088)}          \\ 
& & \small{Label Spreading} & \small{11.457 (0.786)}         \\\cline{2-4} 
& \multirow{3}{*}{Yes} & \small{GM (P=112)}     & \textbf{\small{0.403 (0.044)}} \\
& & \small{kNN} & \small{0.663 (0.036)}          \\ 
& & \small{Label Spreading} & \small{6.500 (0.726)}          \\\hline
\end{tabular}
}
\caption{
\textbf{Graph Dirichlet energy comparison (lower is better).}  
The metric quantifies the average squared gradient magnitude of the class-probability field, where smaller values denote smoother decision transitions.  
\textbf{Graph Memory (GM)} yields substantially lower energy (\textbf{30-40\%} below $k$NN and over \textbf{94\%} below Label Spreading) indicating smoother, more reliable boundaries while preserving accuracy and calibration.
}
\label{tab:smooth}
\end{table}

\subsection{Breast Histopathology (IDC)}
\label{subsec:idc}

\begin{figure}[t]
    \centering
    \includegraphics[width=\linewidth]{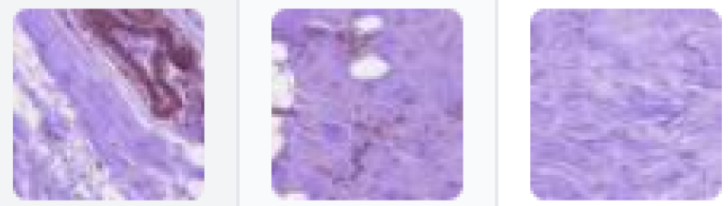}
    \caption{
    Representative benign tissue patches from the IDC dataset.
    }
    \label{fig:idc-benign}
\end{figure}

\begin{figure}[t]
    \centering
    \includegraphics[width=\linewidth]{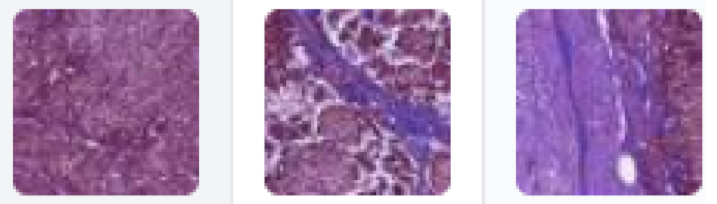}
    \caption{
    Representative malignant tissue patches from the IDC dataset.
    }
    \label{fig:idc-malign}
\end{figure}

We evaluate Graph Memory (GM) on the \textit{Breast Histopathology Images (Invasive Ductal Carcinoma, IDC)} dataset (binary: IDC vs.\ non-IDC)\footnote{Available at \url{https://huggingface.co/datasets/dbzadnen/breast-histopathology-images}}~\cite{janowczyk2016deep,cruz2014automatic}.  
Representative examples of benign and malignant tissue patches are shown in Figs.~\ref{fig:idc-benign} and~\ref{fig:idc-malign}, respectively.
A ResNet18 encoder provides frozen 512-D embeddings.
GM uses $K{=}32$ prototypes with the same hyperparameters as in Sec.~\ref{subsec:synthetic}.
Table~\ref{tab:acc_idc} shows that GM attains the highest accuracy and near-parametric NLL despite using only 32 prototypes.
Table~\ref{tab:idc_smooth} shows that GM also achieves the lowest graph Dirichlet energy, indicating stable, smooth decision functions.

\begin{table}[h]
\begin{tabular}{l|ll}
\hline
Method           & Acc $\uparrow$         & NLL $\downarrow$       \\\hline
GM(P=32) (Ours)     & \textbf{0.845 (0.000)} & 0.420 (0.000)          \\
kNN-budget(P=32) & 0.797 (0.078)          & 0.607 (0.339)          \\
kNN              & 0.840 (0.000)          & 1.039 (0.000)          \\
Label Spreading   & 0.826 (0.000)          & 0.531 (0.000)          \\
Linear           & 0.836 (0.000)          & \textbf{0.397 (0.000)}\\\hline
\end{tabular}
\caption{
\textbf{Accuracy and calibration on the IDC dataset.}
GM achieves the highest accuracy and competitive calibration using only 32 prototypes, demonstrating compact and reliable non-parametric inference.
}
\label{tab:acc_idc}
\end{table}

\begin{table}[]
\begin{tabular}{l|l}
\hline
Method         & graph Dirichlet energy \\\hline
GM (P=32) (Ours)      & 0.291 (0.000)          \\
kNN            & 0.321 (0.000)          \\
Label Spreading & 0.344 (0.000)          \\\hline
\end{tabular}
\caption{
\textbf{Graph Dirichlet energy on the IDC dataset.}
Lower values indicate smoother and more stable decision functions.
Graph Memory (GM) achieves the lowest energy, yielding smoother decision boundaries than $k$NN and Label Spreading while maintaining negligible variance across runs.
}
\label{tab:idc_smooth}
\end{table}

\subsection{Multimodal Experiments}
\label{subsec:multimodal_experiments}

\begin{figure*}[h]
\centering
\includegraphics[width=0.445\linewidth]{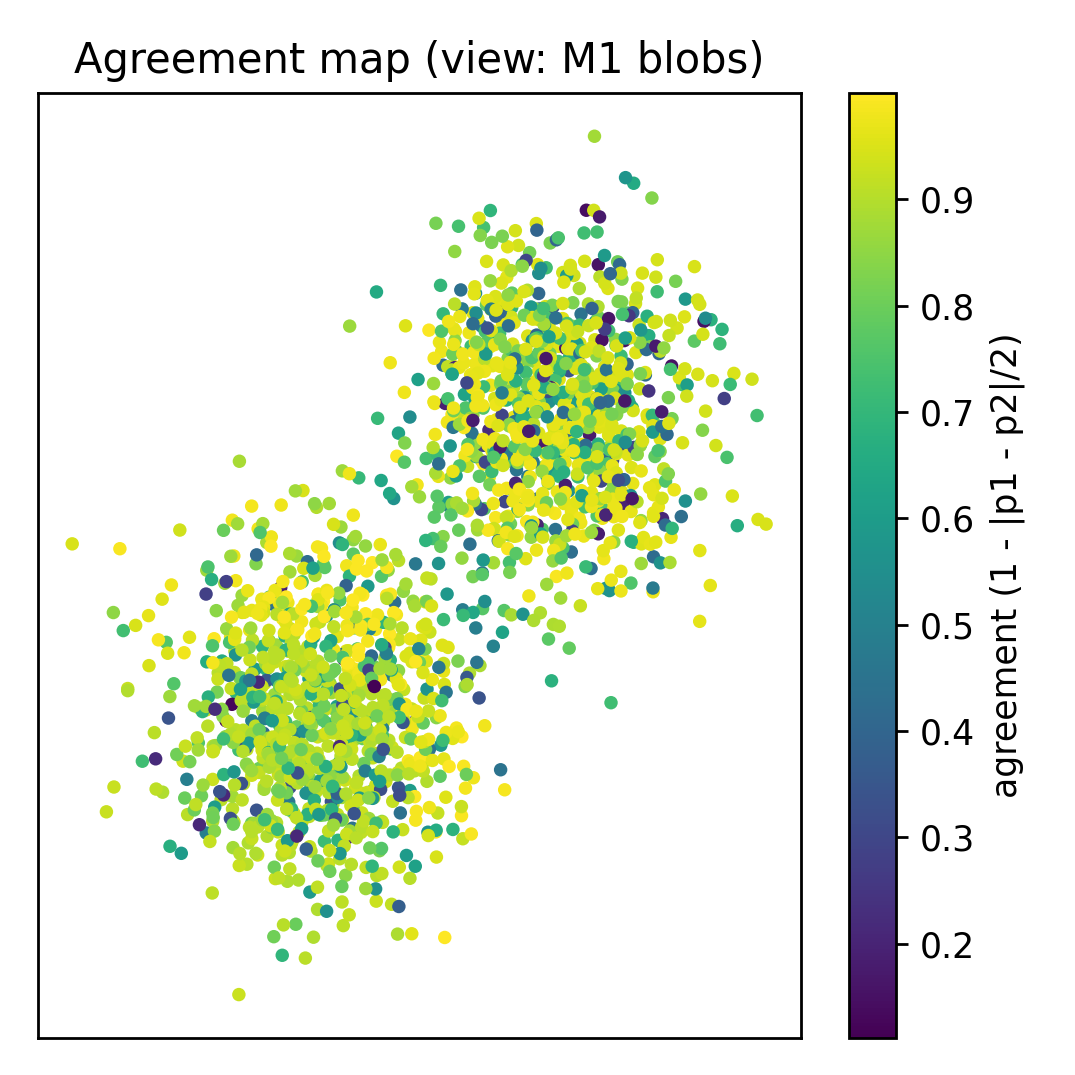}
\includegraphics[width=0.445\linewidth]{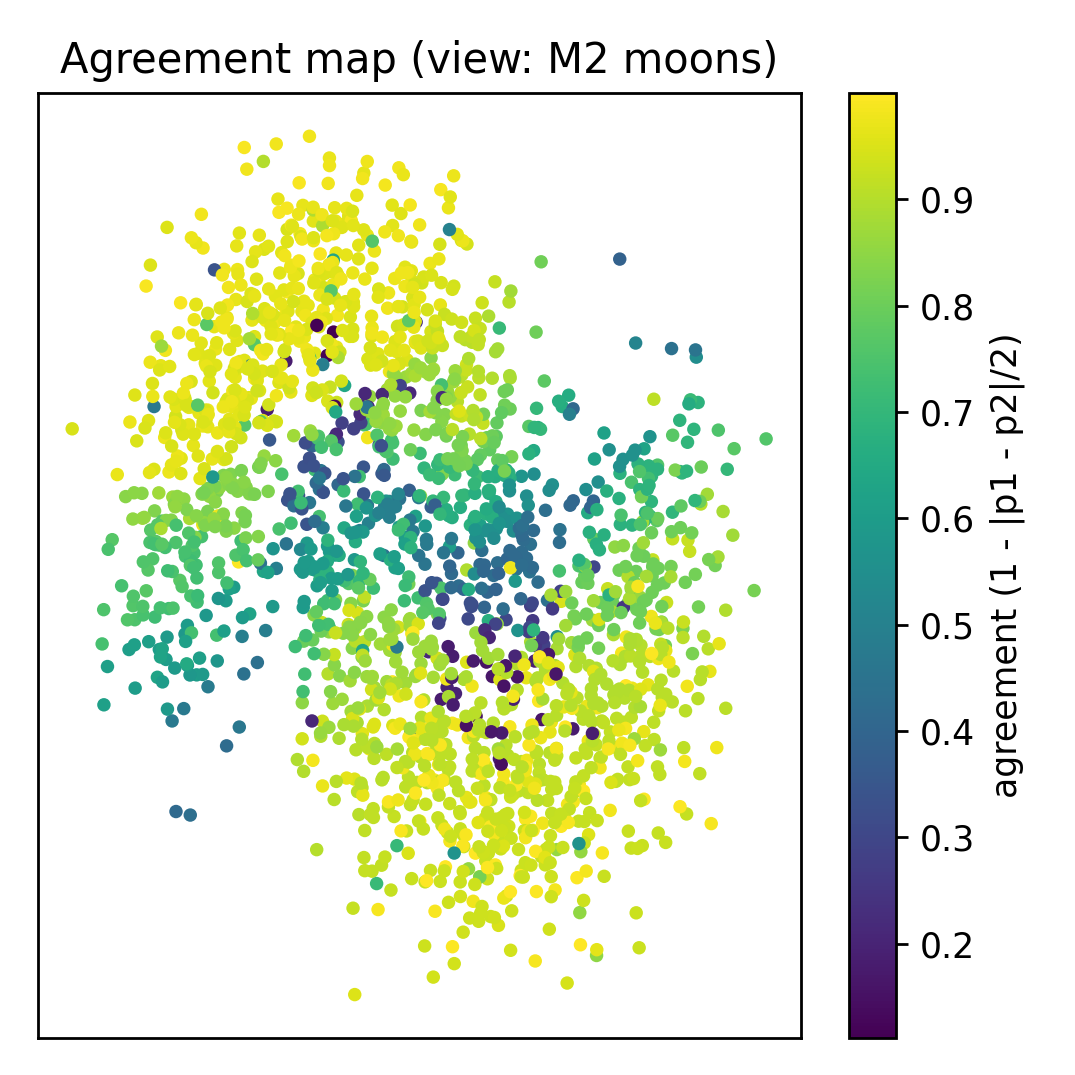}
\caption{
\textbf{Cross-modal agreement maps.}
Agreement ($1 - |p_1 - p_2|/2$) between GM$_1$ (blobs) and GM$_2$ (moons) predictions visualized in each modality’s space.
High agreement (yellow) corresponds to regions where both modalities support the same label with similar confidence, while low agreement (dark) highlights ambiguous or discordant samples.
These visualizations illustrate how independent Graph Memories expose modality complementarity and disagreement without requiring early fusion.
}
\label{fig:multimodal_agreement}
\end{figure*}

\begin{figure*}[h]
    \centering
    \includegraphics[width=0.8\linewidth]{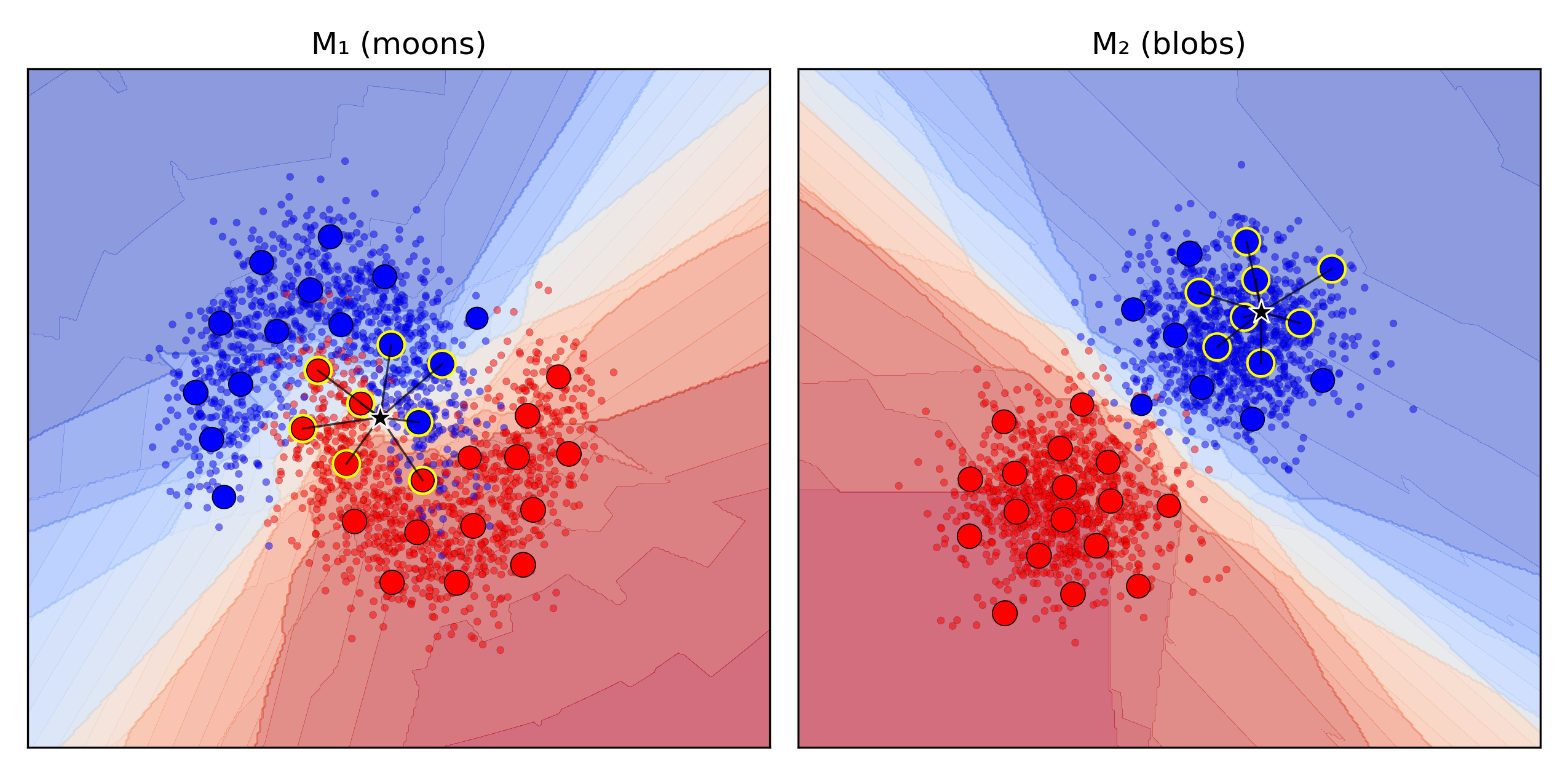}
    \caption{
    \textbf{Prototype-level inference across two complementary modalities.}
    Each panel shows decision regions and reliability-weighted prototypes for one modality: 
    GM$_1$ (\textit{moons}, left) emphasizes non-linear, locally adaptive boundaries, whereas GM$_2$ (\textit{blobs}, right) provides globally consistent but less flexible separation.  
    The query sample ($\bigstar$) connects to a distinct subset of prototypes in each view, illustrating how independent Graph Memories capture modality-specific evidence that can later be combined through reliability-aware fusion.
    }
    \label{fig:multimodal_synth_regions}
\end{figure*}

\paragraph{Synthetic multimodal benchmark.}
We combine two heterogeneous manifolds: \textit{blobs} (globally separable) and \textit{moons} (non-linear).
Independent Graph Memories (GM$_1$, GM$_2$) are built per modality; predictions are fused via reliability-weighted late fusion.
GM$_1$ provides global stability, GM$_2$ fine local adaptation, and fusion yields the best accuracy and smoothest boundaries (Table~\ref{tab:multimodal_synth}, Fig.~\ref{fig:multimodal_synth_regions}).
Cross-modal agreement maps (Fig.~\ref{fig:multimodal_agreement}) reveal complementary modality behavior.
Predictions are combined through reliability-weighted late fusion (Eq.~\ref{eq:fusion_rule}) with equal modality priors ($\omega_1{=}\omega_2{=}1$).

\begin{table}[h]
\centering
\resizebox{\columnwidth}{!}{%
\begin{tabular}{l|ccc}
\hline
Method & Acc $\uparrow$ & NLL $\downarrow$ & E$_{2D}$ $\downarrow$ \\ \hline
GM$_1$ (blobs) & 0.991 (0.002)  & \textbf{0.080 (0.007)} & 0.418 (0.014) \\
GM$_2$ (moons) & 0.873 (0.010)  & 0.325 (0.004) & 0.297 (0.016) \\
GM-fused & \textbf{0.995 (0.
002)} & 0.171 (0.004) & \textbf{0.183 (0.005)} \\ \hline
\end{tabular}
}
\caption{
\textbf{Multimodal synthetic results (blobs + moons).}
The fused Graph Memory (GM-fused) achieves the highest accuracy and the smoothest decision boundaries, combining the global separability of M1 (blobs) with the fine local structure of M2 (moons).
Its reliability-weighted fusion yields more stable, well-calibrated predictions than either single-modality model.
}
\label{tab:multimodal_synth}
\end{table}


\paragraph{Multimodal medical benchmark.}
The AURORA program~\cite{GarciaRecio2023} is a longitudinal effort that profiles paired primary and metastatic breast tumors across multiple institutions. Its RNA-seq component (GSE209998)\footnote{\url{https://www.ncbi.nlm.nih.gov/geo/query/acc.cgi?acc=GSE209998}}~\cite{GarciaRecio2023,Zahraeifard2024,GarciaRecio2025,Edwards2025} provides patient-matched gene-expression data spanning primary, metastatic, and normal reference tissues, with harmonized pipelines and rich clinical annotation.

From a computational standpoint, AURORA presents several challenges: (i) each patient may contribute multiple samples per modality (e.g., several WSI sections or RNA-seq replicates), (ii) modalities have different sample counts, (iii) genomic profiles contain tens of thousands of genes, and (iv) WSI embeddings originate from gigapixel slides. To obtain one standardized representation per modality, all samples belonging to a patient are aggregated via the \textbf{median embedding}.

We embed whole-slide images with TITAN~\cite{ding2024multimodalslidefoundationmodel}. Median aggregation yields a single patient-level descriptor.
Gene-expression profiles are filtered to remove extremely low-variance genes (5510 out of 58{,}388), retaining 52{,}878 informative features. Median aggregation produces one genomic vector per patient.

To obtain a unified latent space, we finetune a \textbf{ResNet18 encoder} independently for each modality. Each network is trained for 20 epochs on a fixed 50/50 split (26/26 patients), replacing the final layer with a 256-D projection head. These embeddings serve as the input to the corresponding Graph Memory.
For each modality we instantiate a Graph Memory with as many prototypes as training samples, using $k_{\text{graph}}{=}6$, attach-$k{=}3$, $\alpha{=}0.1$, and $\beta{=}0.05$. This full-prototype setting avoids compression in the low-$n$ regime. To assess robustness, we repeat training over 10 random seeds and evaluate accuracy, NLL, and graph Dirichlet energy.

Inference is performed independently on each modality. Final predictions use the \textbf{reliability-weighted late fusion} rule from Eq.~\ref{eq:fusion_rule}, combining morphological (WSI) and molecular (RNA-seq) evidence without early-fusion assumptions. This setup mirrors clinical heterogeneity, preserves modality-level interpretability, and enables a controlled comparison of per-modality contributions.
Table~\ref{tab:multimodal_aurora} summarizes accuracy and NLL. Both modalities perform well individually, fusion consistently improves calibration and accuracy. Table~\ref{tab:aurora_smooth} reports the corresponding graph Dirichlet energy, showing that both memories remain smooth and stable despite the extremely small sample size.

\begin{table}[h]
\centering
\begin{tabular}{l|cc}
\hline
Method               & Acc $\uparrow$ & NLL $\downarrow$ \\ \hline
GM$_1$ (WSI)         & 0.933 (0.034)  & 0.917 (0.561)    \\
GM$_2$ (Genetic)     & 0.951 (0.039)  & 0.236 (0.279)    \\
GM-fused             & \textbf{0.971 (0.013)} & \textbf{0.120 (0.025)} \\ \hline
\end{tabular}
\caption{\textbf{AURORA multimodal results.}
Late fusion improves accuracy and calibration by integrating complementary morphological and molecular evidence while preserving per-modality interpretability.}
\label{tab:multimodal_aurora}
\end{table}

\begin{table}[h]
\centering
\begin{tabular}{l|c}
\hline
Method & Dirichlet energy $\downarrow$ \\\hline
GM$_1$ (WSI)     & 0.634 (0.111) \\
GM$_2$ (Genetic) & 0.514 (0.122) \\\hline
\end{tabular}
\caption{\textbf{Graph Dirichlet energy on AURORA.}
Both memories exhibit low energy and modest variance across runs, indicating smooth and stable decision functions even in the small-sample regime.}
\label{tab:aurora_smooth}
\end{table}

\section{\uppercase{Discussion}}

Graph Memory (GM) sits at the intersection of prototype learning, graph-based reasoning, and non-parametric retrieval. The results indicate that explicitly modeling region-level structure provides advantages over traditional instance-level methods such as $k$NN~\cite{knn} and semi-parametric adapters~\cite{papernot2018dknn,doerrich2024privacyknn}. Unlike pure distance-based schemes, GM incorporates reliability estimates and diffusion over a prototype graph, enabling calibrated inference that remains inductive, contrasting with transductive approaches like Label Spreading~\cite{zhu2003ssl,sklearn_labelspreading_archive}.

Classical prototype models (e.g., nearest centroid, Prototypical Networks) provide interpretability but treat prototypes independently. GM extends these ideas by encoding relations between prototypes and propagating evidence through them, bridging prototype abstraction with graph-based consistency. This relational structure explains its advantage over Budget-$k$NN: randomly sampled instances do not capture region-level geometry or reliability.

Graph-based SSL methods~\cite{belkin2006manifold,kipf2017gcn} perform well when the manifold is densely sampled, but often overfit local density and depend on test-time graphs. In contrast, GM operates on region summaries rather than instance graphs, yielding smoother decision functions and better calibration in low-sample regimes such as IDC and the multimodal AURORA dataset.

GM is also inherently interpretable. Prototype reliabilities and edge relations make the contribution of each region explicit, revealing how local support and contextual coherence shape predictions. This aligns with broader trends in structured and multimodal modeling~\cite{oliveira2025deep,oliveira2025creating,oliveira2025improving}, while offering a compact, reusable memory that embeds geometric structure and relational metadata directly into the inference mechanism.

A subtle aspect of GM’s diffusion concerns the restart probability $(1-\alpha)$. As in Personalized PageRank~\cite{haveliwala2002topic}, the restart rate is fixed, while the restart distribution is query-dependent through $z_0$. This yields globally smooth, well-calibrated decision surfaces, though it may dilute local ambiguity when a query lies near unreliable prototypes. An interesting direction is to explore \emph{data-dependent restart schedules}, where $\alpha$ adapts to uncertainty or prototype reliability, potentially improving sensitivity while maintaining inductive diffusion.

Overall, the contrast with non-parametric, prototype-based, and graph-based approaches highlights GM's main strengths: calibrated predictions, smooth decision boundaries, compact memory use, and structured interpretability. These properties make GM a principled and practical alternative in settings where data are scarce, heterogeneous, multimodal, or clinically constrained.

\section{\uppercase{Conclusion}}

This work set out to address a gap in non-parametric inference: the absence of a structured, reusable representation that captures region-level geometry, reliability, and relational context in embedding spaces. We introduced Graph Memory (GM) as a unified framework that combines prototype abstraction, graph-based diffusion, and reliability modeling to produce calibrated, region-aware, and geometrically smooth predictions.

Across synthetic benchmarks, histopathology (IDC), and multimodal biomedical data (AURORA), GM consistently achieved strong accuracy, improved calibration, and lower graph Dirichlet energy compared to classical baselines such as $k$NN, Budget-$k$NN, and Label Spreading. Its compact memory footprint and fully inductive nature make GM suitable for low-sample and high-dimensional regimes, while its explicit node-edge structure supports interpretable, evidence-based reasoning.

By organizing embedding spaces into reliability-annotated prototype graphs, GM provides a principled alternative to instance-based retrieval and conventional graph SSL, extending these paradigms with structured, explainable regional representations. The approach naturally supports multimodal integration through independent memories and reliability-weighted late fusion.

\section*{\uppercase{Acknowledgments}}

This work was funded partially by FAPESP project 2022/15304-4 and MCTI (law 8.248, PPI-Softex - TIC 13 - 01245.010222/2022-44).
The results published here are in whole or part based upon data generated by the TCGA Research Network: \url{https://www.cancer.gov/tcga}.

\bibliographystyle{unsrt}  
\bibliography{references}

\end{document}